\pdfoutput=1

\documentclass[11pt]{article}

\usepackage[]{emnlp2021}

\usepackage{times}
\usepackage{latexsym}

\usepackage[T1]{fontenc}

\usepackage[utf8]{inputenc}

\usepackage{microtype}
\usepackage{amsmath,amsthm}
\usepackage{graphicx}
\usepackage{multirow}
\usepackage{subcaption}
\usepackage{amssymb}
\title{Mitigating Data Scarceness through Data Synthesis, Augmentation and Curriculum for Abstractive Summarization}

\author{Ahmed Magooda \and Diane Litman \\
        University of Pittsburgh \\ Pittsburgh, PA, USA \\ \texttt{\{aem132, dlitman\}@pitt.edu}}

\begin{document}
\maketitle
\begin{abstract}
This paper explores three simple data manipulation techniques (synthesis, augmentation, curriculum) for improving abstractive summarization models without the need for any additional data. We introduce a method of data synthesis with paraphrasing, a data augmentation technique with sample mixing, and curriculum learning with two new difficulty metrics based on specificity and abstractiveness. We conduct experiments to show that these three techniques can help improve abstractive summarization across two summarization models and two different small datasets. Furthermore, we show that these techniques can improve performance when applied in isolation and when combined.
\end{abstract}

\section{Introduction}
Training complex neural models usually requires large amounts of data. However, data annotation still poses a challenge for many domains. Thus, much research focuses on data manipulation (e.g., synthesis, augmentation)  and additional ways to handle data differently during training. 
Prior work on the synthesis of textual data has focused on back translation \cite{parida2019abstract,wang2018switchout,sennrich2016improving} and word replacement~\cite{wang2015s,zhang2015character}. 
We, on the other hand, propose a different approach to data synthesis through paraphrasing. 
However, synthesis  
involves data manipulation on the input level, which  might expose the model to grammatically or logically incorrect input. Thus, 
we  explore a second approach to data manipulation  based on augmentation rather than synthesis. Augmentation aims to move data manipulation from the input side to any part of the model. This, in turn, can help the model be more resilient to over-fitting. Third, we explore using data more efficiently by integrating curriculum learning into the training process. Curriculum learning  reorders training samples based on external criteria, 
which  can help train the model gradually and more efficiently without the need for any external data. We also introduce new difficulty metrics based on specificity and abstractiveness for curriculum construction. Finally, we explore combining multiple techniques (synthesis and curriculum) to overcome the data scarcity issue. Thus, our contribution is threefold: 1) We introduce a simple approach for data synthesis through paraphrasing. 2) We use 
data augmentation by 
sample mixing to move augmentation into 
the model. 3) We integrate a curriculum into the training process and introduce two new difficulty metrics.

\section{Related Work}
\textbf{Abstractive summarization for low resource data}. 
Prior proposed methods for tackling domains with scarce data have included  finetuning pre-trained models \cite{bajaj2021long,yu2021adaptsum,magooda2020abstractive} such as BART \cite{lewis2020bart} 
or using few-shot learning 
\cite{bravzinskas2020few,sarkhel2020interpretable}. Our work differs 
in several aspects. First, our work doesn't focus on improving a certain summarization model; in contrast, we focus on using data efficiently, which can be applied to various models. Second, we focus on techniques that can improve the training process without additional data, e.g.,  synthesis, augmentation, and curriculum learning.

\textbf{Data synthesis and augmentation}. Data synthesis for text summarization is underexplored, with only a few approaches such as back-generation \cite{parida2019abstract} and template-based summary re-writing \cite{magooda2020abstractive}.  We propose doing data synthesis by paraphrasing, which is simpler than the back-translation and template methods. While combining synthesis with paraphrasing has been studied in other contexts \cite{wang2015building,iyyer2018adversarial}, our work differs in both goals and techniques. \citet{wang2015building}  proposed synthesizing data, then crowdsourcing paraphrases to train semantic parsers, while \citet{iyyer2018adversarial}  synthesized data to train a paraphrasing model. Our work, 
to our knowledge, is the first to use a strong language model finetuned for paraphrasing to synthesize data for text summarization.
Finally, for data augmentation, we base our work on the MixText approach  \cite{chen2020mixtext}. 
While the original MixText model is used for classification-based tasks, we introduce a  variation for generative tasks (called MixGen) 
and use it for abstractive summarization.

\textbf{Curriculum learning}. Curriculum learning aims to improve the training procedure with the same amount of data. It has been  applied in NLP \cite{sachan2016easy,sachan2018self,tay2019simple,xu2020curriculum,wang2020learning} for machine comprehension, question generation, reading comprehension, NLU and machine translation, respectively. We build on the approach introduced in \cite{xu2020curriculum}; however, the core differences are both the downstream tasks (classification versus abstractive summarization) and the difficulty metrics. 
In contrast to the only other summarization work that we know of, \citet{kano-etal-2021-quantifying} 
focus on large datasets, while we focus on  low resource domains.  We also introduce two different difficulty metrics (ROUGE and specificity).

\section{Summarization Datasets}

\textbf{CourseMirror  (CM)}\label{data:cmirror}
is a student reflections dataset that has been used in prior work to study extractive \cite{luo2015summarizing} and abstractive \cite{magooda2020abstractive}\footnote{https://petal-cs-pitt.github.io/data.html} summarization. The dataset consists of documents and summaries from four courses. 
\begin{table}[t]
\small
\begin{center}
\begin{tabular}{|l|c|c||c|c|c|}
\hline 
\textbf{Data} & \textbf{\# docs} & \textbf{\# refs} & \textbf{Train} & \textbf{Val} & \textbf{Test}\\
\hline
CM ALL & 368 & 44 & 294 & 37 & 37 \\
\hline
Amazon/Yelp & 160 & 8 & 58 & 42 & 60\\
\hline
\end{tabular}
\end{center}
\caption{\label{dataset_summary} Dataset summary.}
\end{table}
Table \ref{dataset_summary} summarizes the dataset in terms of the \textit{number of documents (\# docs)} and \textit{average reflections per document (\# refs)}. 
We compiled all courses into one dataset (named CM ALL), then split the documents into training, validation, and test sets (80\%, 10\%, 10\%, respectively) by sampling equally from all courses.

\textbf{Amazon/Yelp  (A/Y)} is another small 
 dataset, now consisting of opinions (refer to the appendix for examples) \cite{bravzinskas2020few}\footnote{https://github.com/abrazinskas/FewSum}. 
  The dataset contains customer reviews from Amazon \cite{he2016ups} and Yelp. The data contains 160 products/businesses split into training, validation and test sets as shown in Table \ref{dataset_summary}. Each of the products/businesses contains a set of 8 reviews.

\section{Proposed Model}
\subsection{Synthesis via paraphrasing with GPT-2}
Influenced by  work in style transfer  \cite{krishna2020reformulating}, we propose synthesizing new human summaries by using paraphrasing to generate other potential summaries that are paraphrases of the original human summary. We use the paraphraser trained by \citet{krishna2020reformulating}. They finetuned a large GPT-2 language model with data from PARANMT-50M  \cite{wieting-gimpel-2018-paranmt} to direct the model into generating diverse paraphrases that they later used for style transfer. 

\subsection{Augmentation with sample mixing}
MixText is a data augmentation approach based on mixing two input samples by weight summing the features corresponding to the two samples at any level of the model (specific layer of the encoder, after encoder, etc.) using $\lambda$. The model is then expected to produce a probability distribution over the available classes, similar to a $\lambda$ weighted sum of the two samples' gold predictions. We train the model using KL divergence between the predicted distribution and the expected one. 
We adapt the approach of \citet{chen2020mixtext} for text generation tasks by modifying the decoding process and loss calculation; we  call our approach MixGEN. Like the original MixText, we use two input samples and pass them to the encoder. We pass the samples up to a specific layer, then the two hidden states are summed together weighted differently using the $\lambda$ parameter. On the decoder side, first, we construct the expected values using the following:
\begin{equation*}
\begin{split}
    & for \:i \;\in \;min(L_1, L_2) \;\;PD_{i} = [P_1, P_2, P_3, .. P_v],\\
    & for \:j \;\in \;[1, v] 
    \begin{cases}
    P_j = \lambda, & \text{if } S_1[i]=j\\
    P_j = 1-\lambda, & \text{if } S_2[i]=j\\
    P_j = 0,              & \text{otherwise}
\end{cases}
\\
\end{split}
\end{equation*}
where $v$ = vocab size, and $L_1$, $L_2$ are the human summary of input sample 1 and input sample 2, respectively. $PD_{i}$ is the probability distribution expected for token$_{i}$. $S_1[i]$, $S_2[i]$ are the $i^{th}$ token of the first sample's and second sample's human summary, respectively. In simple wording, during decoding, we expect the output probability distribution across the vocabulary of the decoder at token position $i$ to have two high values, one with value $\lambda$ at vocabulary token corresponding to the $i^{th}$ token of the first sample's human summary, and another value of $1-\lambda$ at vocabulary token corresponding to the $i^{th}$ token of the second sample's human summary. This should continue as long as $i$ is less than or equal to both human summaries' length. Once $i$ is greater than the minimum summaries' length, then the expected distribution would only correspond to the longer summary. Finally, the text generation on the decoder side is auto-regressive, thus, expected token is passed at the end of each generation step. However, If we pass the argmax of the expected distribution, then we will end up always passing the token corresponding to the sample with higher weight (Alpha) vs. (1-Alpha). Thus, we randomly sample from the two input samples based on their weights using the following equation.
\begin{equation*}
\begin{split}
    & for \:i \;\in \;[1,L_{min}], \;P_i \thicksim U(0, 1)\\
    & for \:i \;\in \;[1,L_{min}]
    \begin{cases}
    P_i <= \lambda, & T_i\; from \;S_1\\
    P_i > \lambda, & T_i \;from \;S_2\\
\end{cases}
\\
\end{split}
\end{equation*}

where $S_1$ and $S_2$ are the first and second sample respectively, $L_{min}$ is minimum length of both $S_1$ and $S_2$. $P_i$ is the sampling probability and $U(0,1)$ is a uniform distribution.
\subsection{Curriculum learning (Cur.)}
Curriculum learning aims to help the model training process by introducing easier samples first followed by more difficult ones according to a particular difficulty metric. We use the curriculum construction approach introduced in \citet{xu2020curriculum}. In this approach, we split data into $N$ buckets based on a difficulty metric. We then train the model in a difficulty incremental setting. 
In this work, we use two different curriculum difficulty metrics. 

\textbf{Specificity (S)} measures how specific or vague a piece of text is. We argue that the more specific a piece of text is, the more complicated it can get. For example, text like (Nothing, Everything is Easy, etc.) are not specific and easy for the model to learn and vice versa. 
We feed the model less specific pieces of text first during  training,  then introduce the more specific ones as training progresses. Specificity is calculated (Appendix) on the reflection/review level, so we use the average  values of the whole set of reflections/reviews as the document  value. For example; for a training sample of an input document $D$ consists of $N$ independent reflections/reviews $[r_1, r_2, r_3, .., r_n]$, we calculate the specificity value for the sample as follows:
\begin{equation*}
    D_{s} = \sum_{i=1}^{N} S(r_i)/N
\end{equation*}
where $S(r_i)$ is specificity value of the i's reflection/review.

\textbf{ROUGE (R)} is the standard metric for evaluating summarization performance. Thus, we decided to use ROUGE scores as a difficulty metric. For a training sample, we calculate different ROUGE scores between the input document $D$ and its corresponding human summary $S$, then use \textit{average of (R1, R2, RL)} as the difficulty metric. 
According to \cite{liu2018generating}, the higher the ROUGE score, the less abstractive the summary is compared to the input, and vice versa. We argue that the more abstractive samples are harder to learn.

\section{Experiments}
\subsection{Parameters}
\textbf{Baselines}: To our knowledge, in prior work there is no data synthesis technique used for summarization except back generation \cite{parida2019abstract} and template synthesis \cite{magooda2020abstractive}. Thus, we developed two  synthesis baselines \textbf{(shuffle; shuffle + mask)}. We generated 10 
samples for each of the original training samples for both 
baselines by randomly shuffling the reflections/reviews. Additionally, for the shuffle-mask baseline, we randomly mask 50\% of the reflection/reviews 50\% of the time.

\textbf{Paraphrasing  with  GPT-2}: We generate $N$ synthetic samples for each original sample by generating $N$ paraphrases of the human summary and shuffle the input reflections. We varied N between [5, 10] to monitor the effect of synthetic data size.

\textbf{MixGEN}: We integrate MixGEN by combining each sample with $N$ other samples during training. We used $N$=3 for our experiments. 
Moreover, we use mixing probability $\alpha$=0.75 as specified by the original code implementation\footnote{https://github.com/GT-SALT/MixText}.  

\textbf{Curriculum learning:} In the curriculum learning experiments, we use a specificity prediction model that consists of a DistilBERT \cite{sanh2019distilbert} encoder with a logistic regression classification layer (Appendix). We normalize the whole training data values between 1 and $N$, where $N$ is the number of buckets to split the data. We use $N$=10. Similarly, we normalize the average ROUGE value to also be between 1 and $N$.

\subsection{Model Training}
In all of our experiments, we use the BERTSum\footnote{Easy to use and one of the SOTA summarization models.\\https://github.com/nlpyang/PreSumm} 
 model proposed by \citet{liu2019text}. We used the same parameters in the original code (Appendix). We conducted experiments on  CM and A/Y datasets using proposed methods  in a regular training  and in a \textbf{(pretraining$\rightarrow$fine-tuning)} setting, where we perform pretraining with synthesized data and fine-tuneing using original data.

\section{Results}
\begin{table}[ht]
\begin{center}
\small
\begin{tabular}{|l|l|c|c|c|}
\hline
  \multicolumn{2}{|c|}{}  & \multicolumn{3}{|c|}{\bf CM} \\
\hline 
Pretraining & finetuning &  R1 &  R2 & RL \\
\hline
\multicolumn{5}{|c|}{No Pretraining}\\
\hline

\multirow{6}{*}{None}   & Original  & 36.34 & 11.39 & 26\\
\cline{2-5}

& shuff.  & \bf 38.57 & 11.72 & 26.94\\
\cline{2-5}

& shuff.+mask  & 37.07 & 11.52 & 26.45\\
\cline{2-5}
& Cur.(S)  & 36.88 & \bf 12.41 & \bf 27.63\\
\cline{2-5}

& Cur.(R) & 37.01 & 12.13 & 27.11\\
\cline{2-5}

& Mix(n=3)  & 36.87 & 11.98 & 26.57\\
\hline
\hline
\multicolumn{5}{|c|}{With synthetic data pretraining}\\
\hline

\multirow{3}{*}{Synth.(n=5)}  & Original  & 39.39 & 12.85 & 26.66\\
\cline{2-5}

& Cur.(S) & 40.68 & 13.59 & 26.26\\
\cline{2-5}

& Cur.(R) & 39.35 & 12.33 & 26.48\\
\hline

\multirow{3}{*}{Synth.(n=10)}  & Original  & \bf 41.14 & \bf 14.24 & 26.98\\
\cline{2-5}

& Cur.(S)  & 39.81 & 12.94 & 26.81\\
\cline{2-5}

& Cur.(R)  & 40.22  & 14.12  & \bf 27.33\\
\hline

\end{tabular}
\end{center}
\caption{\label{tab:bertsum_results_CM} ROUGE results of BERTSum model with different augmentation techniques on CM data.}
\end{table}

\begin{table}[!ht]
\begin{center}
\small
\begin{tabular}{|l|l|c|c|c|}
\hline
  \multicolumn{2}{|c|}{} & \multicolumn{3}{|c|}{\bf A/Y} \\
\hline 
Pretraining & finetuning &  R1 &  R2 &  RL \\
\hline
\multicolumn{5}{|c|}{No Pretraining}\\
\hline

\multirow{6}{*}{None}   & Original  & 27.71 & 3.83 & 17.83\\
\cline{2-5}

& shuff.  & 28.34 & 4.04 & 17.74\\
\cline{2-5}

& shuff.+mask  & 28.01 & 4.21 & 17.87\\
\cline{2-5}
& Cur.(S)  & 28.69 & 4.28 & 17.95\\
\cline{2-5}

& Cur.(R) & \bf 28.8 & \bf 4.33 & \bf 18.18\\
\cline{2-5}

& Mix(n=3)  & 27.85 & 3.95 & 17.89\\
\hline
\hline
\multicolumn{5}{|c|}{With synthetic data pretraining}\\
\hline

\multirow{3}{*}{Synth.(n=5)}  & Original  & 28.27 & 4.36 & 17.84\\
\cline{2-5}

& Cur.(S) & 27.95 & 4.4 & 18.01\\
\cline{2-5}

& Cur.(R) & \bf 28.56 & 4.25 & 18.06\\
\hline

\multirow{3}{*}{Synth.(n=10)}  & Original  & 28.49 & \bf 4.54 & 18.08\\
\cline{2-5}

& Cur.(S)  & 28.52 & 4.5 & \bf 18.15\\
\cline{2-5}

& Cur.(R)  & 28.21 & 4.35 & 17.87\\
\hline

\end{tabular}
\end{center}
\caption{\label{tab:bertsum_results_Amazon} 
ROUGE results of BERTSum model with different augmentation techniques on A/Y data.}
\end{table}

Tables \ref{tab:bertsum_results_CM} and \ref{tab:bertsum_results_Amazon} show results obtained through conducting experiments on CM and A/Y datasets.  Considering \textbf{data  synthesis and augmentation}, we first see that the two baselines (shuffle and shuffle+mask) can improve performance compared to no data manipulation across all ROUGE scores except RL for shuffle baseline on the A/Y dataset. This shows that reducing the model dependency on the input sentence order can help the model depend more on the actual input text. 
Moving to the proposed augmentation technique (MixGEN), we see that we can get a performance gain across all ROUGE scores across both datasets by mixing training samples compared to normal training with a single sample. Similarly, we can see that providing synthetic data with the proposed paraphrasing approach can help outperform both using original data as well as baselines with (41.14, 14.24, 26.98) compared to (36.34, 11.39, 26) and (38.57, 11.72, 26.94) for original and shuffle baseline respectively on CM, and (28.49, 4.54, 18.08) compared to (27.71, 3.83, 17.83) and (28.34, 4.04, 17.74) on A/Y. Additionally, we can see that increasing the synthetic data size helps to improve the model performance across all ROUGE scores for both CM and A/Y datasets (N=5 vs. N=10).

Now moving to \textbf{curriculum learning}, we can see that integrating a curriculum to reorder training data differently using any of the two proposed difficulty metrics can lead to consistent improvements across all ROUGE scores for both CM and A/Y datasets. Additionally, we can see that curriculum can improve scores compared to the two augmentation baselines across all ROUGE scores except R1 for CM data. On the other hand, we don't see consistent ROUGE score improvement when using curriculum for fine-tuning after pretraining with synthetic data. We hypothesize that this behavior might be due to performing the pretraining phase without curriculum integration, unlike the fine-tuning phase. We plan to conduct experiments with a curriculum integrated into both pretraining and fine-tuning to validate our hypothesis. Furthermore, while both curriculum difficulty metrics (i.e., Specificity and ROUGE) introduced improvement compared to training with no curriculum, we didn't observe any consistent improvement pattern in using one metric over the other.

\section{Conclusion and Future work}
In this work, we showed that we could mitigate the effect of data scarcity in different datasets (i.e., CourseMirror and Amazon/Yelp) for abstractive summarization using three simple data manipulation techniques. We showed that synthesizing data with paraphrasing to use for pretraining can boost the model performance across all ROUGE scores for different datasets. Additionally, we showed that mixing samples for training can also push the model to be more resilient to overfitting and improve its performance. Finally, we showed that reordering training samples through curriculum, using the proposed difficulty metrics \textbf{(i.e., Specificity, and ROUGE)} would help improve all ROUGE scores across different datasets without the need for any additional data (either true or synthetic). In the future, we plan to try more $N$ values for synthesis and MixGen. Additionally, we plan to investigate other curriculum difficulty metrics. We plan to use BART model as one of the SOTA models for abstractive summarization. Finally, we are doing additional experiments on multitask learning, and we plan to combine both techniques in one framework targeting low resource domains.

\section*{Acknowledgements}
The research reported here was supported, in whole or in
part, by the institute of Education Sciences, U.S. Department
of Education, through Grant R305A180477 to the University of Pittsburgh. The opinions expressed are those of the
authors and do not represent the views of the institute or the
U.S. Department of Education. We like to thank the Pitt PETAL group and the anonymous reviewers for advice in improving this paper.


\bibliography{CurSumArxiv}

\begin{thebibliography}{28}
\expandafter\ifx\csname natexlab\endcsname\relax\def\natexlab#1{#1}\fi

\bibitem[{Bajaj et~al.(2021)Bajaj, Dangati, Krishna, Ashok~Kumar, Uppaal,
  Windsor, Brenner, Dotterrer, Das, and McCallum}]{bajaj2021long}
Ahsaas Bajaj, Pavitra Dangati, Kalpesh Krishna, Pradhiksha Ashok~Kumar, Rheeya
  Uppaal, Bradford Windsor, Eliot Brenner, Dominic Dotterrer, Rajarshi Das, and
  Andrew McCallum. 2021.
\newblock \href {https://doi.org/10.18653/v1/2021.acl-srw.7} {Long document
  summarization in a low resource setting using pretrained language models}.
\newblock In \emph{Proceedings of the 59th Annual Meeting of the Association
  for Computational Linguistics and the 11th International Joint Conference on
  Natural Language Processing: Student Research Workshop}, pages 71--80,
  Online. Association for Computational Linguistics.

\bibitem[{Bra{\v{z}}inskas et~al.(2020)Bra{\v{z}}inskas, Lapata, and
  Titov}]{bravzinskas2020few}
Arthur Bra{\v{z}}inskas, Mirella Lapata, and Ivan Titov. 2020.
\newblock \href {https://doi.org/10.18653/v1/2020.emnlp-main.337} {Few-shot
  learning for opinion summarization}.
\newblock In \emph{Proceedings of the 2020 Conference on Empirical Methods in
  Natural Language Processing (EMNLP)}, pages 4119--4135, Online. Association
  for Computational Linguistics.

\bibitem[{Chen et~al.(2020)Chen, Yang, and Yang}]{chen2020mixtext}
Jiaao Chen, Zichao Yang, and Diyi Yang. 2020.
\newblock \href {https://doi.org/10.18653/v1/2020.acl-main.194} {{M}ix{T}ext:
  Linguistically-informed interpolation of hidden space for semi-supervised
  text classification}.
\newblock In \emph{Proceedings of the 58th Annual Meeting of the Association
  for Computational Linguistics}, pages 2147--2157, Online. Association for
  Computational Linguistics.

\bibitem[{Fan et~al.(2017)Fan, Luo, Menekse, Litman, and Wang}]{fan2017scaling}
Xiangmin Fan, Wencan Luo, Muhsin Menekse, Diane Litman, and Jingtao Wang. 2017.
\newblock Scaling reflection prompts in large classrooms via mobile interfaces
  and natural language processing.
\newblock In \emph{Proceedings of the 22nd International Conference on
  Intelligent User Interfaces}, pages 363--374. ACM.

\bibitem[{He and McAuley(2016)}]{he2016ups}
Ruining He and Julian~J. McAuley. 2016.
\newblock \href {https://doi.org/10.1145/2872427.2883037} {Ups and downs:
  Modeling the visual evolution of fashion trends with one-class collaborative
  filtering}.
\newblock In \emph{Proceedings of the 25th International Conference on World
  Wide Web, {WWW} 2016, Montreal, Canada, April 11 - 15, 2016}, pages 507--517.
  {ACM}.

\bibitem[{Iyyer et~al.(2018)Iyyer, Wieting, Gimpel, and
  Zettlemoyer}]{iyyer2018adversarial}
Mohit Iyyer, John Wieting, Kevin Gimpel, and Luke Zettlemoyer. 2018.
\newblock \href {https://doi.org/10.18653/v1/N18-1170} {Adversarial example
  generation with syntactically controlled paraphrase networks}.
\newblock In \emph{Proceedings of the 2018 Conference of the North {A}merican
  Chapter of the Association for Computational Linguistics: Human Language
  Technologies, Volume 1 (Long Papers)}, pages 1875--1885, New Orleans,
  Louisiana. Association for Computational Linguistics.

\bibitem[{Kano et~al.(2021)Kano, Takahashi, Nishino, Taniguchi, Taniguchi, and
  Ohkuma}]{kano-etal-2021-quantifying}
Ryuji Kano, Takumi Takahashi, Toru Nishino, Motoki Taniguchi, Tomoki Taniguchi,
  and Tomoko Ohkuma. 2021.
\newblock \href {https://aclanthology.org/2021.eacl-main.119} {Quantifying
  appropriateness of summarization data for curriculum learning}.
\newblock In \emph{Proceedings of the 16th Conference of the European Chapter
  of the Association for Computational Linguistics: Main Volume}, pages
  1395--1405, Online. Association for Computational Linguistics.

\bibitem[{Krishna et~al.(2020)Krishna, Wieting, and
  Iyyer}]{krishna2020reformulating}
Kalpesh Krishna, John Wieting, and Mohit Iyyer. 2020.
\newblock \href {https://doi.org/10.18653/v1/2020.emnlp-main.55} {Reformulating
  unsupervised style transfer as paraphrase generation}.
\newblock In \emph{Proceedings of the 2020 Conference on Empirical Methods in
  Natural Language Processing (EMNLP)}, pages 737--762, Online. Association for
  Computational Linguistics.

\bibitem[{Lewis et~al.(2020)Lewis, Liu, Goyal, Ghazvininejad, Mohamed, Levy,
  Stoyanov, and Zettlemoyer}]{lewis2020bart}
Mike Lewis, Yinhan Liu, Naman Goyal, Marjan Ghazvininejad, Abdelrahman Mohamed,
  Omer Levy, Veselin Stoyanov, and Luke Zettlemoyer. 2020.
\newblock \href {https://doi.org/10.18653/v1/2020.acl-main.703} {{BART}:
  Denoising sequence-to-sequence pre-training for natural language generation,
  translation, and comprehension}.
\newblock In \emph{Proceedings of the 58th Annual Meeting of the Association
  for Computational Linguistics}, pages 7871--7880, Online. Association for
  Computational Linguistics.

\bibitem[{Liu et~al.(2018)Liu, Saleh, Pot, Goodrich, Sepassi, Kaiser, and
  Shazeer}]{liu2018generating}
Peter~J. Liu, Mohammad Saleh, Etienne Pot, Ben Goodrich, Ryan Sepassi, Lukasz
  Kaiser, and Noam Shazeer. 2018.
\newblock \href {https://openreview.net/forum?id=Hyg0vbWC-} {Generating
  wikipedia by summarizing long sequences}.
\newblock In \emph{6th International Conference on Learning Representations,
  {ICLR} 2018, Vancouver, BC, Canada, April 30 - May 3, 2018, Conference Track
  Proceedings}. OpenReview.net.

\bibitem[{Liu and Lapata(2019)}]{liu2019text}
Yang Liu and Mirella Lapata. 2019.
\newblock \href {https://doi.org/10.18653/v1/D19-1387} {Text summarization with
  pretrained encoders}.
\newblock In \emph{Proceedings of the 2019 Conference on Empirical Methods in
  Natural Language Processing and the 9th International Joint Conference on
  Natural Language Processing (EMNLP-IJCNLP)}, pages 3730--3740, Hong Kong,
  China. Association for Computational Linguistics.

\bibitem[{Luo and Litman(2015)}]{luo2015summarizing}
Wencan Luo and Diane Litman. 2015.
\newblock \href {https://doi.org/10.18653/v1/D15-1227} {Summarizing student
  responses to reflection prompts}.
\newblock In \emph{Proceedings of the 2015 Conference on Empirical Methods in
  Natural Language Processing}, pages 1955--1960, Lisbon, Portugal. Association
  for Computational Linguistics.

\bibitem[{Magooda and Litman(2020)}]{magooda2020abstractive}
Ahmed Magooda and Diane Litman. 2020.
\newblock Abstractive summarization for low resource data using domain transfer
  and data synthesis.
\newblock In \emph{The Thirty-Third International Flairs Conference}.

\bibitem[{Parida and Motlicek(2019)}]{parida2019abstract}
Shantipriya Parida and Petr Motlicek. 2019.
\newblock \href {https://doi.org/10.18653/v1/D19-1616} {Abstract text
  summarization: A low resource challenge}.
\newblock In \emph{Proceedings of the 2019 Conference on Empirical Methods in
  Natural Language Processing and the 9th International Joint Conference on
  Natural Language Processing (EMNLP-IJCNLP)}, pages 5994--5998, Hong Kong,
  China. Association for Computational Linguistics.

\bibitem[{Sachan and Xing(2016)}]{sachan2016easy}
Mrinmaya Sachan and Eric Xing. 2016.
\newblock \href {https://doi.org/10.18653/v1/P16-1043} {Easy questions first? a
  case study on curriculum learning for question answering}.
\newblock In \emph{Proceedings of the 54th Annual Meeting of the Association
  for Computational Linguistics (Volume 1: Long Papers)}, pages 453--463,
  Berlin, Germany. Association for Computational Linguistics.

\bibitem[{Sachan and Xing(2018)}]{sachan2018self}
Mrinmaya Sachan and Eric Xing. 2018.
\newblock \href {https://doi.org/10.18653/v1/N18-1058} {Self-training for
  jointly learning to ask and answer questions}.
\newblock In \emph{Proceedings of the 2018 Conference of the North {A}merican
  Chapter of the Association for Computational Linguistics: Human Language
  Technologies, Volume 1 (Long Papers)}, pages 629--640, New Orleans,
  Louisiana. Association for Computational Linguistics.

\bibitem[{Sanh et~al.(2019)Sanh, Debut, Chaumond, and
  Wolf}]{sanh2019distilbert}
Victor Sanh, Lysandre Debut, Julien Chaumond, and Thomas Wolf. 2019.
\newblock \href {https://arxiv.org/abs/1910.01108} {Distilbert, a distilled
  version of bert: smaller, faster, cheaper and lighter}.
\newblock \emph{ArXiv preprint}, abs/1910.01108.

\bibitem[{Sarkhel et~al.(2020)Sarkhel, Keymanesh, Nandi, and
  Parthasarathy}]{sarkhel2020interpretable}
Ritesh Sarkhel, Moniba Keymanesh, Arnab Nandi, and Srinivasan Parthasarathy.
  2020.
\newblock \href {https://doi.org/10.18653/v1/2020.coling-main.606}
  {Interpretable multi-headed attention for abstractive summarization at
  controllable lengths}.
\newblock In \emph{Proceedings of the 28th International Conference on
  Computational Linguistics}, pages 6871--6882, Barcelona, Spain (Online).
  International Committee on Computational Linguistics.

\bibitem[{Sennrich et~al.(2016)Sennrich, Haddow, and
  Birch}]{sennrich2016improving}
Rico Sennrich, Barry Haddow, and Alexandra Birch. 2016.
\newblock \href {https://doi.org/10.18653/v1/P16-1009} {Improving neural
  machine translation models with monolingual data}.
\newblock In \emph{Proceedings of the 54th Annual Meeting of the Association
  for Computational Linguistics (Volume 1: Long Papers)}, pages 86--96, Berlin,
  Germany. Association for Computational Linguistics.

\bibitem[{Tay et~al.(2019)Tay, Wang, Luu, Fu, Phan, Yuan, Rao, Hui, and
  Zhang}]{tay2019simple}
Yi~Tay, Shuohang Wang, Anh~Tuan Luu, Jie Fu, Minh~C. Phan, Xingdi Yuan, Jinfeng
  Rao, Siu~Cheung Hui, and Aston Zhang. 2019.
\newblock \href {https://doi.org/10.18653/v1/P19-1486} {Simple and effective
  curriculum pointer-generator networks for reading comprehension over long
  narratives}.
\newblock In \emph{Proceedings of the 57th Annual Meeting of the Association
  for Computational Linguistics}, pages 4922--4931, Florence, Italy.
  Association for Computational Linguistics.

\bibitem[{Wang et~al.(2020)Wang, Tian, Ngiam, Yang, Caswell, and
  Parekh}]{wang2020learning}
Wei Wang, Ye~Tian, Jiquan Ngiam, Yinfei Yang, Isaac Caswell, and Zarana Parekh.
  2020.
\newblock \href {https://doi.org/10.18653/v1/2020.acl-main.689} {Learning a
  multi-domain curriculum for neural machine translation}.
\newblock In \emph{Proceedings of the 58th Annual Meeting of the Association
  for Computational Linguistics}, pages 7711--7723, Online. Association for
  Computational Linguistics.

\bibitem[{Wang and Yang(2015)}]{wang2015s}
William~Yang Wang and Diyi Yang. 2015.
\newblock \href {https://doi.org/10.18653/v1/D15-1306} {That{'}s so
  annoying!!!: A lexical and frame-semantic embedding based data augmentation
  approach to automatic categorization of annoying behaviors using {\#}petpeeve
  tweets}.
\newblock In \emph{Proceedings of the 2015 Conference on Empirical Methods in
  Natural Language Processing}, pages 2557--2563, Lisbon, Portugal. Association
  for Computational Linguistics.

\bibitem[{Wang et~al.(2018)Wang, Pham, Dai, and Neubig}]{wang2018switchout}
Xinyi Wang, Hieu Pham, Zihang Dai, and Graham Neubig. 2018.
\newblock \href {https://doi.org/10.18653/v1/D18-1100} {{S}witch{O}ut: an
  efficient data augmentation algorithm for neural machine translation}.
\newblock In \emph{Proceedings of the 2018 Conference on Empirical Methods in
  Natural Language Processing}, pages 856--861, Brussels, Belgium. Association
  for Computational Linguistics.

\bibitem[{Wang et~al.(2015)Wang, Berant, and Liang}]{wang2015building}
Yushi Wang, Jonathan Berant, and Percy Liang. 2015.
\newblock \href {https://doi.org/10.3115/v1/P15-1129} {Building a semantic
  parser overnight}.
\newblock In \emph{Proceedings of the 53rd Annual Meeting of the Association
  for Computational Linguistics and the 7th International Joint Conference on
  Natural Language Processing (Volume 1: Long Papers)}, pages 1332--1342,
  Beijing, China. Association for Computational Linguistics.

\bibitem[{Wieting and Gimpel(2018)}]{wieting-gimpel-2018-paranmt}
John Wieting and Kevin Gimpel. 2018.
\newblock \href {https://doi.org/10.18653/v1/P18-1042} {{P}ara{NMT}-50{M}:
  Pushing the limits of paraphrastic sentence embeddings with millions of
  machine translations}.
\newblock In \emph{Proceedings of the 56th Annual Meeting of the Association
  for Computational Linguistics (Volume 1: Long Papers)}, pages 451--462,
  Melbourne, Australia. Association for Computational Linguistics.

\bibitem[{Xu et~al.(2020)Xu, Zhang, Mao, Wang, Xie, and
  Zhang}]{xu2020curriculum}
Benfeng Xu, Licheng Zhang, Zhendong Mao, Quan Wang, Hongtao Xie, and Yongdong
  Zhang. 2020.
\newblock \href {https://doi.org/10.18653/v1/2020.acl-main.542} {Curriculum
  learning for natural language understanding}.
\newblock In \emph{Proceedings of the 58th Annual Meeting of the Association
  for Computational Linguistics}, pages 6095--6104, Online. Association for
  Computational Linguistics.

\bibitem[{Yu et~al.(2021)Yu, Liu, and Fung}]{yu2021adaptsum}
Tiezheng Yu, Zihan Liu, and Pascale Fung. 2021.
\newblock \href {https://doi.org/10.18653/v1/2021.naacl-main.471}
  {{A}dapt{S}um: Towards low-resource domain adaptation for abstractive
  summarization}.
\newblock In \emph{Proceedings of the 2021 Conference of the North American
  Chapter of the Association for Computational Linguistics: Human Language
  Technologies}, pages 5892--5904, Online. Association for Computational
  Linguistics.

\bibitem[{Zhang et~al.(2015)Zhang, Zhao, and LeCun}]{zhang2015character}
Xiang Zhang, Junbo~Jake Zhao, and Yann LeCun. 2015.
\newblock \href
  {https://proceedings.neurips.cc/paper/2015/hash/250cf8b51c773f3f8dc8b4be867a9a02-Abstract.html}
  {Character-level convolutional networks for text classification}.
\newblock In \emph{Advances in Neural Information Processing Systems 28: Annual
  Conference on Neural Information Processing Systems 2015, December 7-12,
  2015, Montreal, Quebec, Canada}, pages 649--657.

\end{thebibliography}
\bibliographystyle{acl_natbib}

\appendix
\section{BERTSum training parameters}
We train the model for 200K steps using a batch size of 140. We use 20K steps for BERT warmup, 10K steps for decoder warmup, and a max position of 512. We use 4 Nvidia Quadro RTX 5000 GPUs. We use the checkpoint with highest ROUGE score on validation set for testing.

\section{MixGen Model}
\begin{figure}[!ht]
\centering
\begin{subfigure}{0.48\textwidth}
  \centering
  \includegraphics[width=.95\linewidth]{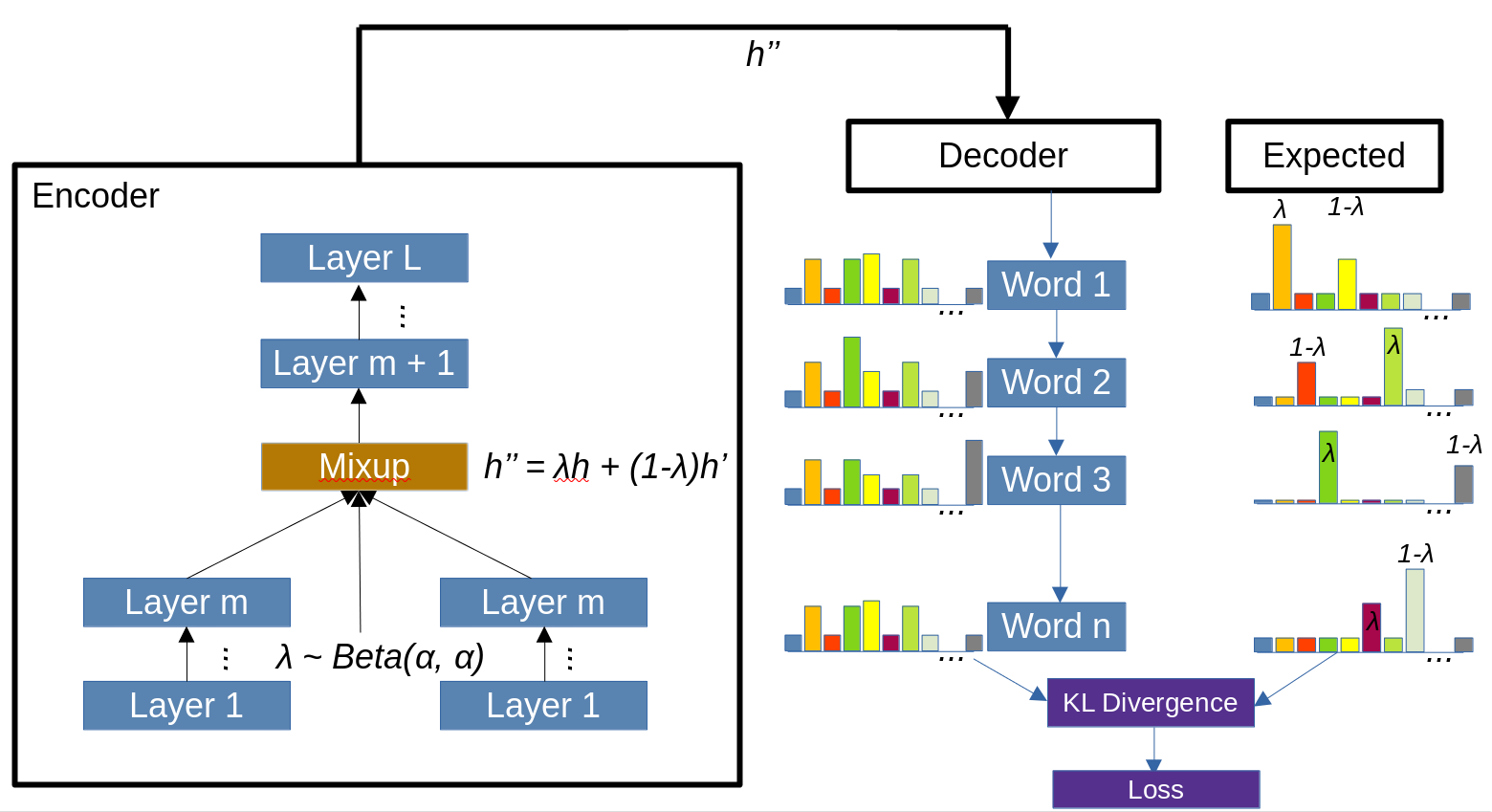}
  \caption{MixText For generative tasks.}
  \label{fig:generative_mixtext}
\end{subfigure}\\
\begin{subfigure}{.48\textwidth}
  \centering
  \includegraphics[width=.95\linewidth]{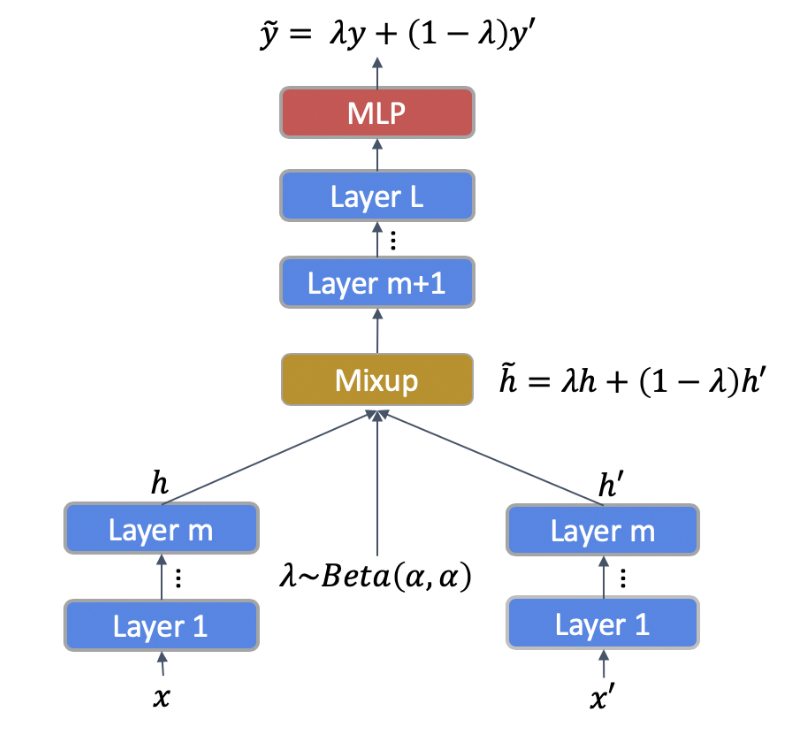}
  \caption{Original MixText.}
  \label{fig:orig_mixtext}
\end{subfigure}
\caption{Mixtext model and the modified MixGen for generative tasks.}
\label{fig:MixText}
\end{figure}

Figure \ref{fig:MixText} shows both the original MixText model and the modified MixText for generative tasks (MixGen). 

\section{Specificity Model}
\label{sec:appendix_1}

\subsection{Model}
\begin{figure}[!htpb]
 \centering
    \includegraphics[width=0.48\textwidth]{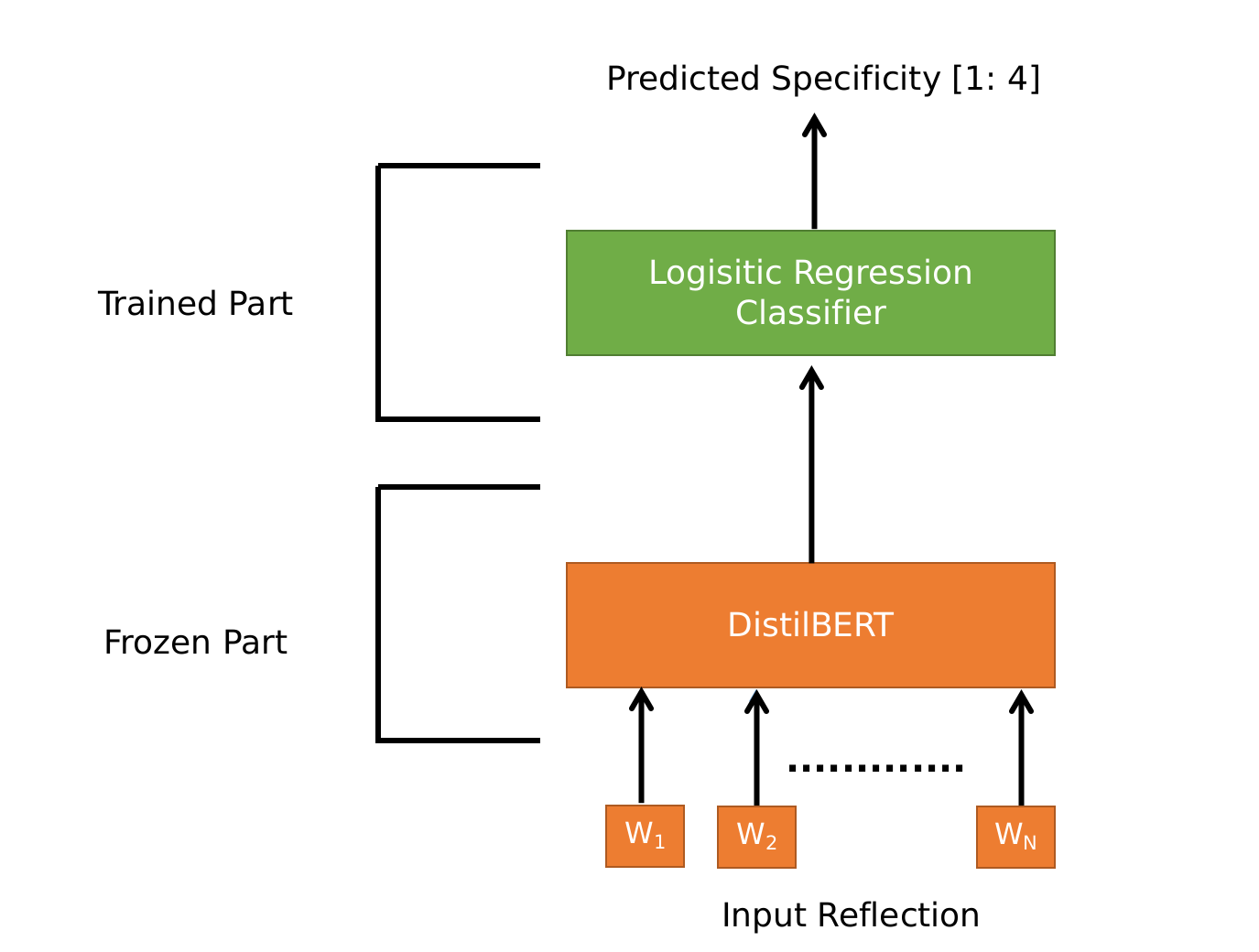}
    \caption{Specificity prediction model used.}
    \label{fig:spec_model_fig}
\end{figure}

CourseMirror data is also annotated for specificity. The data contains human annotations for around 7000 reflections\footnote{https://petal-cs-pitt.github.io/data.html} using the scheme introduced in \cite{fan2017scaling}. Table \ref{specificity_summary} shows the score distribution for CourseMirror specificity dataset. We use the data to train a specificity predicition model. We use the model to predict the specificity values for both CourseMirror and Amazon/Yelp datasets. The specificity prediction model (figure \ref{fig:spec_model_fig}) uses DistilBERT encoder to produce reflection embedding, the embeddings are then used as features to train a logistic regression classifier. To keep the number of tuned parameters to minimum, the DistilBERT weights are frozen during the training process. The embeddings are used as fixed features, and all the training is performed on the logistic classifier side.

\begin{table}[!ht]

\begin{center}
\begin{tabular}{|c|c|c|c|}
\hline 
\textbf{1} & \textbf{2} & \textbf{3} & \textbf{4}\\
\hline
1354 & 2035 & 2377 & 1058\\
\hline
\end{tabular}
\end{center}
\caption{\label{specificity_summary} CourseMirror Specificity dataset score distribution.}
\end{table}

\section{Data samples}
\subsection{CourseMirror (CM)}
Table \ref{tab:cmirror_summaries_example} shows an example of CM sample from CS course.
\begin{table*}[htpb]
\small
\begin{tabular}{|p{0.96\textwidth}|}
\hline \small\textbf{Prompt}\\
\hline \small Point of Interest (POI): Describe what you found most interesting in today's class.\\
\hline \small\textbf{Student Reflection Document}\\
\hline

\textbullet $ $ the dynamic bag\\
\textbullet $ $ I found the creation of the Bag to be the most interesting.\\
\textbullet $ $ Learning about bags was very interesting.\\
\textbullet $ $ Dr. Ramirez cleared up my understanding of how they should work.\\
\textbullet $ $ I was really interested in learning all about an entirely new data structure , the Bag.\\
\textbullet $ $ I 'm also noticing that as these classes get farther along , there is more focus on real world factors that determine strength of code like speed\\
\textbullet $ $ The bag concept was cool how basically acts like a bag in real life with its usefulness.\\
\textbullet $ $ Bags as a data type and how flexible they are.\\
\textbullet $ $ Discussing the Assignment 1\\
\textbullet $ $ I found the examples and drawings the teacher drew on the whiteboard the most interesting.\\
\textbullet $ $ Abstraction, though seemingly intimidating is kind of just giving programmers a break right?\\
\textbullet $ $ We 're given so many more abilities and operations without having to know exactly how to code that.\\
\textbullet $ $ That being said , while I understand the applications being explained to me , it 's hard to just manifest that on my own.\\
\textbullet $ $ Learning about resizing Bags dynamically\\
\textbullet $ $ The discussion of the underlying methods of ADTs such as bags was most interesting\\
\textbullet $ $ the implementation of an array bag\\
\textbullet $ $ Order does not matter when using a bag.\\
\textbullet $ $ It is important to keep all of the values in an array together.\\
\textbullet $ $ To do this , you should move an existing element into the vacant spot.\\
\textbullet $ $ Looking at ADT 's from both perspectives\\
\textbullet $ $ Information held in bags is not in any particular order\\
\textbullet $ $ different ways to implement the bag\\
\textbullet $ $ Thinking about a more general idea of coding with ADTs and starting to dig into data structures more specifically.\\
\textbullet $ $ Code examples of key concepts/methods is always helpful.\\
\textbullet $ $ I thought it was a good thing to go through the implementation of both the add ( ) and remove ( ) methods of the Bag ADT\\
\textbullet $ $ Today we were talking about a certain type of ADT called a bag.\\
\textbullet $ $ We talked about certain ways that we would implement the methods and certain special cases that we as programmers have to be aware of.\\
\textbullet $ $ If you were removing items from ADT bag , you can simply shift the bottom or last item and put it in the place where you we removed an item.\\
\textbullet $ $ This is because , in bags , order does not matter.\\
\textbullet $ $ Learning about managing arrays in a data structure\\
\textbullet $ $ The bag ADT and how it is implemented\\

\hline \textbf{Reference Abstractive Summary}\\\hline
 Students were interested in ADT Bag, and also its array implementation. Many recognized that it should be resizable, and that the underlying array organization should support that. Others saw that order does not matter in bags. Some thought methods that the bag provides were interesting.\\
\hline
\end{tabular}
\caption{\label{tab:cmirror_summaries_example} Sample data from the CourseMirror CS course.}
\end{table*} 

\subsection{Amazon/Yelp}
Table \ref{tab:amazon_summaries_example} shows an example of sample from amazon/Yelp data.
\begin{table*}[htpb]
\small
\begin{tabular}{|p{0.96\textwidth}|}
\hline
\bf Reviews\\
\hline
This pendant is so unique!! The design is beautiful and the bail is a ring instead of the typical bail which gives it a nice touch!! All the corners are smooth and my daughter loves it - looks great on her.I cannot say anything about the chain because used our own chain.:) Satisfied.\\

\hline  
It look perfect in a womens neck!! great gift, I thought for the price it was going to look cheap, but I was far wrong. It look great.Spect great reward from your woman when you give this to her; D\\

\hline
 The prettiest sterling silver piece I own now. I get so many compliments on this necklace. I bought it for myself from my hubby for Valentine's Day. Why not? When people ask where I got it, I simply say from my loving hubby. And he is off the hook as to what to get me. win + win.\\

\hline
I love hearts and I love 'love':) I do not have any negative feedback, the necklace is perfect and the charm is perfect. I just thought it would have been slightly bigger. Overall, I love my new heart necklace.\\

\hline
When I received the package, I was surprised and amazed because the necklace is so elegant, beautiful and the same as the picture shown here. I really love this necklace. It has a unique pendant designed. I will recommend it to someone to order it now...\\

\hline
Item is nice. Not a great quality item, but right for the price. Charm was larger than I expected (I expected small and elegant, but it was large and almost costume jewelry like). I think it is a good necklace, just not what I expected.\\

\hline
I got this as a present for my GF on Valintines day. She loves it and wears it every day! Its not cheap looking and it hasn't broken yet. The chain hasn't broken either even though it is very thin. Strongly recomend it!\\

\hline
Over all service has been great the only problem, I ordered a purple Mickey Mouse case for iPhone 4S they sent a black, n I felt it was to much trouble n such a small item to send back so needless to say its put back in a drawer somewhere\\
\hline
\hline
\bf Abstractive Summary\\
\hline
This silver chain and pendant are elegant and unique. The necklace is very well made, making it a great buy for the cost, and is of high enough quality to be worn every day. The necklace looks beautiful when worn bringing many compliments. Overall, it is highly recommended.\\
\hline
\end{tabular}
\caption{\label{tab:amazon_summaries_example} Sample data from the Amazon/Yelp data.}
\end{table*}

\end{document}